\documentclass[letterpaper, 10 pt, conference]{ieeeconf}

\usepackage{graphicx}
\usepackage{subfigure}
\usepackage{bm}
\usepackage{amsfonts}
\usepackage{amssymb}
\usepackage{multirow}
\usepackage{amsmath}
\usepackage{cite}

\IEEEoverridecommandlockouts
\title{\LARGE \bf
Understand Scene Categories by Objects: A Semantic Regularized Scene Classifier Using Convolutional Neural Networks}

\author{Yiyi Liao$^{1}$, Sarath Kodagoda$^{2}$, Yue Wang$^{1}$, Lei Shi$^{2}$, Yong Liu$^{1}$ 
\thanks{$^{1}$Yiyi Liao, Yue Wang and Yong Liu are with the State Key Laboratory of Industrial Control Technology and Institute of Cyber-Systems and Control, Institute of Cyber-Systems and Control, Zhejiang University, Zhejiang, 310027, China.}
\thanks{$^{2}$Sarath Kodagoda and Lei Shi are with the Centre for Autonomous Systems (CAS), The University of Technology, Sydney, Australia.}}

\begin{document}
\maketitle

\begin{abstract}

Scene classification is a fundamental perception task for environmental understanding in today's robotics. In this paper, we have attempted to exploit the use of popular machine learning technique of deep learning to enhance scene understanding, particularly in robotics applications. As scene images have larger diversity than the iconic object images, it is more challenging for deep learning methods to automatically learn features from scene images with less samples. Inspired by human scene understanding based on object knowledge, we address the problem of scene classification by encouraging deep neural networks to incorporate object-level information. This is implemented with a regularization of semantic segmentation. With only 5 thousand training images, as opposed to 2.5 million images, we show the proposed deep architecture achieves superior scene classification results to the state-of-the-art on a publicly available SUN RGB-D dataset. In addition, performance of semantic segmentation, the regularizer, also reaches a new record with refinement derived from predicted scene labels. Finally, we apply our SUN RGB-D dataset trained model to a mobile robot captured images to classify scenes in our university demonstrating the generalization ability of the proposed algorithm.


\end{abstract}

\section{Introduction}

\begin{figure}[tpb]
\centering
\includegraphics[height=6cm]{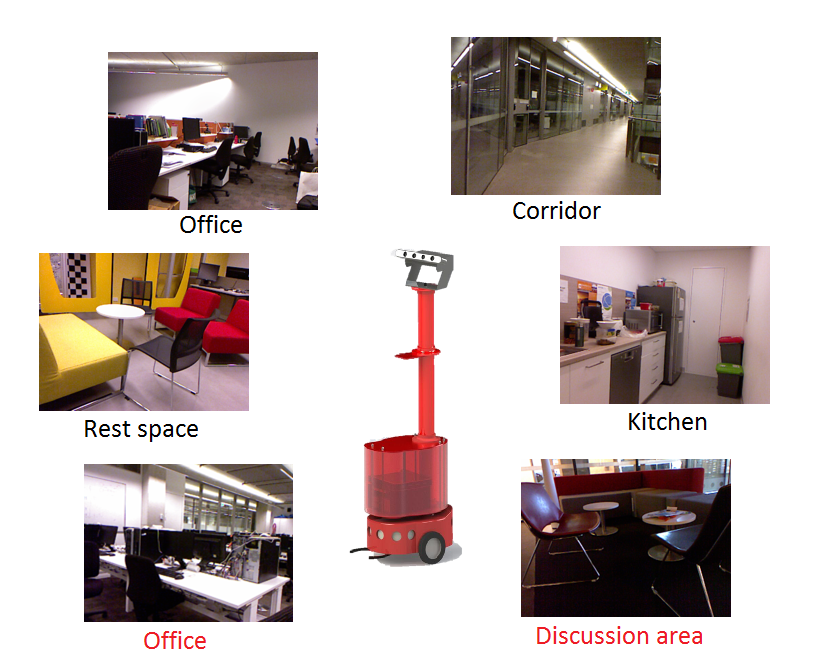}
\caption{Scene classification demonstration. The examples are captured in our university using the mobile robot. Our SS-CNN trained on SUN RGB-D dataset gives the predict labels below each image, without retraining for the completely new environment. Labels in black means correct classification. Two misclassified image are given with labels in red, while the predicted labels are in accord with human recognition to some extent.}
\label{robot}
\end{figure}

Today's robotics face many perception challenges such as scene classification (Figure~\ref{robot}), semantic segmentation, object recognition and detection. For object-level tasks, a series of new performance standards are set with the recently successful deep Convolutional Neural Networks (CNN)~\cite{krizhevsky2012imagenet,donahue2013decaf,razavian2014cnn}, while the performance on scene-level perception based on deep CNN did not reach the same level of success before the work of Place-CNN~\cite{zhou2014learning}.
As pointed out in~\cite{zhou2014learning}, scene-level task is more challenging for feature learning due to the larger diversity of scene images compared to iconic object images. For Place-CNN, it overcame this diversity and reached state-of-the-art by training on 2.5 million scene images. However, it is very expensive to collect and label the training images in such a large scale. Furthermore, enhancing the performance by increasing the number of training samples is not preferable in most robotic applications, especially for those tasks with insufficient samples. In this paper, we focus on constructing a scene classifier with competitive performance, while automatically learns feature with less amount of training images using the deep CNN.

\begin{figure*}[tpb]
\centering
\includegraphics[height=6cm]{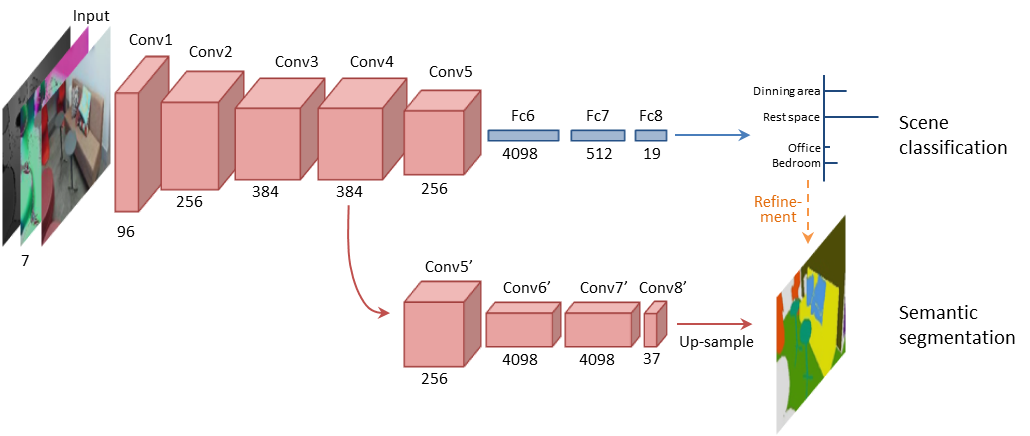}
\caption{An example architecture of the proposed SS-CNN, which is composed of a main branch for scene classification and a regularizer branch for semantic segmentation. The semantic regularization is imposed to the beginning 4 layers in this figure. The main branch outputs 1-D probability prediction for each image, where the regularization branched outputs 2-D probability prediction for each pixel. A refinement process denoted as the dashed orange line is implemented in the test process to promote the performance of the semantic segmentation using the predicted scene labels.}
\label{fig_framework}
\end{figure*}

It is more likely that the human beings understand the scene classes mainly according to the object-level information, as scene classes are naturally defined at a higher level than the objects.
For example, we incline to recognize the scene as ``bedroom'' as we find the objects ``bed'' and ``night stand'' in it.
Intuitively, understanding the scene classes involving object-level information would suppress the large diversity on scene images and lead to better generalization ability.
This hypothesis is validated with a preliminary experiment by using object existence as feature vector to classify the scene classes --- with a much lower dimension, the object existence feature allows a similar performance to the Place-CNN features. This result reveals that object-level information has the potential to improve scene classification.
Inspired by human way of scene classification, we encourage the deep CNN to understand objects in early stage. Specifically,
we develop a \textit{scene classification} model with regularization of \textit{semantic segmentation} based on the well-known CNN architecture, Alexnet~\cite{krizhevsky2012imagenet}, named SS-CNN. An example of our model structure is shown in Figure~\ref{fig_framework}, where the features learned for scene classification in SS-CNN automatically involves object-level information. On SUN RGB-D dataset~\cite{Song2015}, we train our SS-CNN and show it significantly outperforms the original Alexnet, which further validate our hypothesis that the semantic regularization enhances the generalization ability. Besides, SS-CNN achieves superior results compared to the state-of-the-art Place-CNN, which is also based on Alexnet but gains its power with 2.5 million training images, while SS-CNN is only trained with 5 thousand images.

In addition, we develop a refinement method for semantic segmentation with the predicted scene classes, which is based on scene-object co-occurrences learned from training data. For instance, knowing the scene as a ``bedroom'' could prevent us from misclassifying the object ``cabinet'' as a ``fridge''.  As a result, the performance of semantic segmentation also reaches the state-of-the-art on SUN RGB-D dataset.

After training and validation on SUN RGB-D dataset collected in the US, we further apply our SS-CNN to a mobile robot to classify scenes in a building of our university, in Australia. The mobile robot and some RGB images with its predicted results are given in Figure~\ref{robot}. The promising performance of SS-CNN in the completely new environment reveals its potential capability in robotics applications.

The remainder of the paper is organized as follow: Section~\ref{sec_related} gives a review of related works and Section~\ref{sec_explory} gives the preliminary experiment to validate our hypothesis.  The proposed model and refinement method are given in Section~\ref{sec_model}, and the experimental results are shown in Section~\ref{sec_exp}. Finally, we conclude the paper with future direction of research in Section~\ref{sec_conclu}.

\section{Related works}\label{sec_related}

Some previous works have demonstrated that interaction between scene and objects have the capability to promote each other\cite{yao2012describing,lin2013holistic,luo2011simultaneous,rogers2012conditional}. The typical idea is to build the relationship between scenes and objects using a graphical model such as Markov Random Field or Conditional Random Field~\cite{lafferty2001conditional}. Though these works have achieved superior results, they are based on hand-crafted features, which means the feature extraction and classification in these works are not in a unified optimization framework.
Compared to these works focusing on simultaneously labeling, our work is more close to Object Bank~\cite{li2010object} since we focus on regularizing scene classification with semantic segmentation. Object Bank proposed a high-level representation for scene classification by encoding the images with combination of a large amount of object detectors. However, the feature extraction and scene classification in Object Bank are still optimized separately, and it requires pre-training a large number of object detectors. Recently, the superior results achieved with deep learning methods suggest that learning features with a fully trainable architecture may be a better choice. In this paper, we implemented the scene classifier using a fully trainable deep architecture with a single semantic segmentation branch encoding all object-level information.

As for the conventional deep learning methods, the most successful CNN model in scene classification is Place-CNN~\cite{zhou2014learning}, which is trained on 2.5 million labeled images belonging to 476 scene classes using the well-known architecture Alexnet~\cite{krizhevsky2012imagenet}. Before~\cite{zhou2014learning}, the performance of CNN on scene classification was within the range of performances of some hand-crafted features based implementations. As pointed out in~\cite{zhou2014learning}, one reason of the relatively poor result on scene classification of CNN is due to the larger diversity of scene-centric images compared to object-centric images, which means scene classification has higher requirement on generalization ability. By encouraging CNN to classify the scene through implicit understanding of object existence, we developed a scene classifier also based on Alexnet and achieves better generalization ability than Place-CNN with only 5 thousand training images.

To the best of our knowledge, considering multiple tasks is rarely exploited in deep learning methods. An exception is the refinement in DeepID-Net~\cite{ouyang2015deepid}, which took the image classification result to refine the object detection. More specifically, they introduced another separated network for image classification and concatenated the estimated image probability with the estimated object probability for a further classification, which means the information of two tasks are only combined after independent training, instead of simultaneous training as implemented in this paper.

\section{Exploratory study}\label{sec_explory}

To explore the role that object-level information plays in scene classification, we first conduct a preliminary experiment to classify a scene with objects occurrence knowledge. We assume the ground truth of object occurrences is known in every image, then each image can be represented by a binary encoded vector with length $M_o$, with 1 denotes the object is contained in the image and 0 otherwise, and $M_o$ is the number of object classes in the whole dataset.

We test this approach on the SUN RGB-D dataset~\cite{Song2015}, a recently published indoor dataset with dense annotations for both objects and scenes. Here the coarse annotation of semantic segmentation is adopted where $M_o=37$. We follow the same split configuration as in~\cite{Song2015} to test the performance and Table~\ref{table_nb} shows the scene classification results based on different features. A brief introduction of comparing methods are given in Section~\ref{sec_exp}. Here both GIST and Place-CNN features are extracted from RGB image.

\begin{table}[tpb]
\caption{Scene classification accuracy based on object occurrence with comparison to benchmark.}
\label{table_nb}
\begin{center}
\begin{tabular}{lc}
Method   											& Acc (\%)  \\
\hline
GIST + RBF Kernel SVM~\cite{Song2015} 				& 19.7		\\
Place-CNN + Linear SVM~\cite{Song2015}				& 35.6 		\\
Place-CNN + RBF Kernel SVM~\cite{Song2015}			& 38.1		\\
Object occurrence + Linear SVM 	 					& 33.1 		\\
\end{tabular}
\end{center}
\end{table}

Experimental results show that object occurrences significantly outperforms the hand-crafted feature GIST, and reaches a similar level to the Place-CNN. It is to be noted that the dimension of object occurrence feature is much lower than both features of GIST and Place-CNN. This experiment reveals that the knowledge on object level have the potential to promote the performances of scene classification, which inspires us to consider learning scene classification involved with object-level information.

\section{Model design}\label{sec_model}

With the aim of learning scene features involved object information, we construct our SS-CNN for \textit{scene classification} with regularization of \textit{semantic segmentation}. In this section, the network architecture of our SS-CNN is introduced in detail, followed by the model learning and input construction. On top of that, we implement refinement for semantic segmentation with the predicted scene labels.

\begin{figure*}[tpb]
\centering
\subfigure[SS-CNN-R6]{\label{ss_cc_r6}
\includegraphics[height=3.58cm]{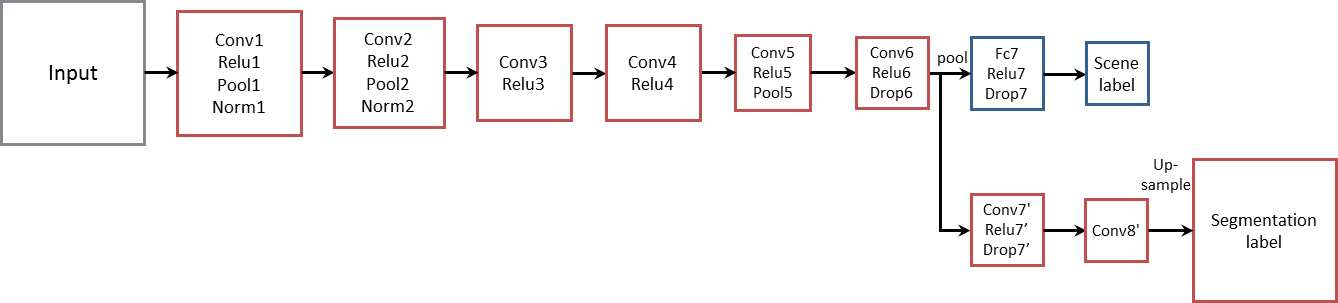}}
\subfigure[SS-CNN-R8]{\label{ss_cc_r8}
\includegraphics[height=3.53cm]{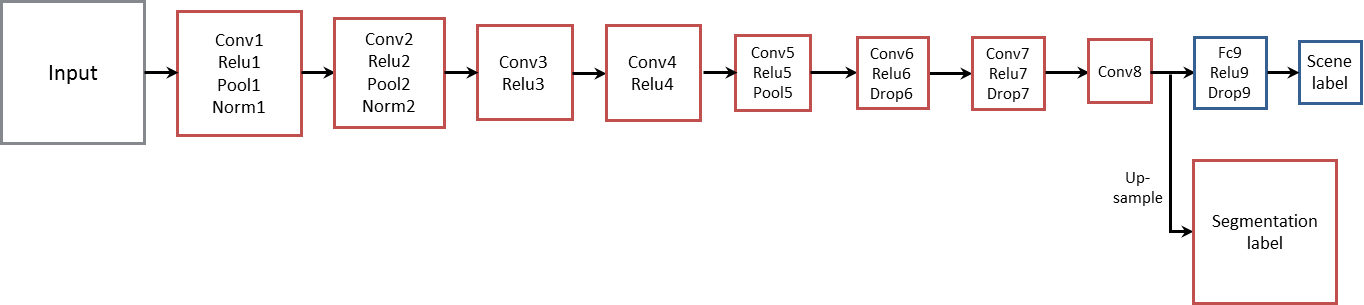}}
\caption{Examples of SS-CNN-R$n$ with $n=6,8$. Note that the structure in Figure~\ref{fig_framework} is SS-CNN-R4. The main branch in SS-CNN-R4 has the same structure with Alexnet with 5 convolutional layers and 3 fully connected layers, while SS-CNN-R6 have 6 convolutional layers and 2 fully connected layers. SS-CNN-R8 is more special with its 8 convolutional layers and 2 additional fully connected layers in the main branch.}
\label{fig_ss_cnn}
\end{figure*}

\subsection{CNN for scene classification with semantic segmentation regularization}\label{sec_model1}

\textbf{Notation.} We first clarify the symbols used in this paper. Assume there are $M_s$ scene classes in scene classification and $M_o$ object classes in semantic segmentation. Let's denote the data structure of a single sample as $(\bm{X},\bm{y_s},\bm{Y_o})$, where $\bm{X}\in \mathbb{R}^{H \times W \times C}$ is input image with $H$ as height, $W$ as width and $C$ as number of channels, $\bm{y_s}\in \mathbb{Z}^{1 \times M_s}$ is the ground truth of a scene label encoded in 1-of-K encoding scheme, i.e. $y_s^k=1$ if $\bm{X}$ belongs to $k^{th}$ scene class, otherwise $y_s^k=0$. $\bm{Y_o} \in \mathbb{Z}^{H \times W \times M_o}$ is the ground truth of semantic segmentation label having the same height and width with $\bm{X}$. Analogously, $y_o^{ijk} = 1$ denotes the pixel $(i,j)$ belonging to $k^{th}$ object class.

\textbf{Network architecture.}  For model construction of SS-CNN, a conventional CNN model is employed as the basic model to predict the scene classes with input pair $(\bm{X},\bm{y_s})$. Then the major contribution of this paper is to add another fully convolutional branch~\cite{long2014fully} to the basic model, with the aim of estimating $\bm{Y_o}$ for semantic segmentation. The fully convolutional branch can be added to the main branch on arbitrary layer, we further define SS-CNN-R$n$ to denote the different configurations of SS-CNN as follow:

\textit{Given an original CNN for scene classification with $N_l$ layers in all, denote SS-CNN-R$n$ as the SS-CNN with the previous $n$ layers regularized by semantic segmentation, $n$ is ranging from 1 to $N_l$. }


In this paper, we take the well-known Alexnet architecture~\cite{krizhevsky2012imagenet} as our main branch for scene classification. In Alexnet, we have $N_l=8$ and there are 8 invariants of SS-CNN. The detailed network configuration of some typical networks are given in Figure~\ref{fig_ss_cnn}.

Intuitively, how many layers are regularized by semantic segmentation would influence the performance of SS-CNN. If $n$ is small, then the regularization is only imposed to a few layers of the scene classification. Considering the extreme case with $n=0$, then two separate neural networks are constructed for scene classification and semantic segmentation respectively. As $n$ getting larger, the semantic segmentation regularizes more layers in the main branch.

It is to be noted that the main branch keeps its original structure from SS-CNN-R1 to SS-CNN-R5 with 5 convolutional layers and 3 fully connected layers. Beginning from SS-CNN-R6, the structure of the main branch is slightly different, as the fully connected layers in main branch are also casted into convolutional layers one by one. When $n=8$, \texttt{fc6} and \texttt{fc7} are both casted into convolutional layer, thus two additional fully connected layers are built for scene classification.

\textbf{Model learning.} As can be seen from the SS-CNN architectures, the loss function of our SS-CNN is composed of two parts, one is the loss of scene classification and the other is the semantic segmentation. In this paper, we use the multinomial logistic loss on a softmax layer. The loss function of scene classification is:
\[ L_{scene}=-\sum_{k=1}^{M_s} y_s^k log(p_s^k)
\]
where $p_s^k$ is the probability of estimating $\bm{X}$ in class $k$, which is obtained with the final softmax layer taking $\bm{f}$ as input:
\begin{equation}\label{eq_prob}
p_s^k = \frac{e^{\bm{f}^T \bm{\theta_k} }}{\sum_{i=1}^{N_s}{e^{\bm{f}^T \bm{\theta_i} }}}
\end{equation}
Analogously, we can obtain the probability of each pixel $p_o^{ijk}$ in semantic segmentation branch and define the loss function as:
\[ L_{object}=-\sum_i \sum_j \sum_{k=1}^{M_o} y_o^{ijk}log(p_o^{ijk})
\]
Then the loss of the whole network is composed of these two losses as:
\[ L_{ss} = L_{scene} + \alpha  L_{object}
\]
where $\alpha$ is the weight of the regularization term $L_{object}$. Notice that each image is corresponding to a single cost for scene classification, while the cost of semantic segmentation is summarized over all pixels (not normalized in our model), thus we take a fixed weight $\alpha=1/1000$ to balance this two costs in this paper.

We use stochastic gradient descent with momentum for model training. Note that given SS-CNN-R$n$, only weights from layer 1 to layer $n$ are regularized with semantic segmentation, i.e. tuned with respect to the partial gradient of $L_{ss}$. From layer $n+1$, the weights in scene classification branch is tuned with only respect to $L_{scene}$, and the same for the semantic segmentation branch as being tuned with respect to $\alpha L_{object}$.

\textbf{Depth representation.} Depth information is important in scene understanding. Many successful models are built on RGB-D inputs captured by the affordable RGB-D sensors such as Kinect and X-tion, especially in indoor environments~\cite{couprie2013indoor,gupta2014learning}. In this paper, we also explore the effective ways to encode the depth information in deep CNN.

\begin{figure}[tpb]
\centering
\subfigure[RGB Image]{
\includegraphics[height=2cm]{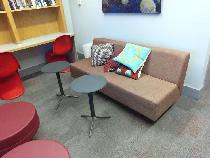}}
\subfigure[Depth Image]{
\includegraphics[height=2cm]{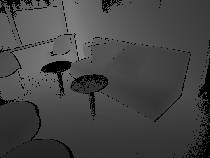}}
\subfigure[Normal Vector]{
\includegraphics[height=2cm]{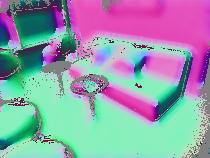}}
\caption{An example of the RGB image in SUN RGB-D dataset, with its corresponding depth image and normal vector image. }
\label{fig_example}
\end{figure}

The most direct way of considering depth information in deep CNN is to add a depth channel in the input layer. The depth image we use is linearly rescaled to $[0,255]$, which is in the same range as the RGB image. Since depth image only provides information of distances, we also consider using the knowledge of normal vectors. For estimation of normal vectors, the depth image is first applied to a bilateral filter for smoothing. And then the smoothed depth image is transformed into a point cloud with the camera intrinsic parameters, on which normal vector is estimated. The normal vector is also rescaled to $[0,255]$ and represented in an image with 3 channels. A visualization of the RGB image and its corresponding depth image and normal vector is given in Figure~\ref{fig_example}.

In this paper, we encode the depth representation as a combination of depth image and normal vector image, and then the RGB-D input has 7 channels for each image as shown in Figure~\ref{fig_framework}.

\subsection{Refinement of semantic segmentation with scene classification}

Intuitively, scene classes can provide prior information about object occurrences. This idea could be used to further refine the performance of the semantic segmentation. For example, if a robot recognizes an environment correctly as a bedroom, then it is fair to expect a bed in the image, rather than a shower curtain. Based on the architecture of SS-CNN, we can conveniently incorporate the estimated scene probability to refine the performance of semantic segmentation.

\begin{figure}[tpb]
\centering
\includegraphics[height=4cm]{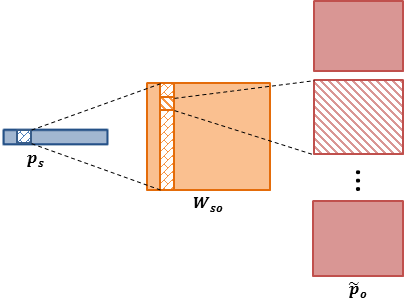}
\caption{Illustration of refinement process.}
\label{fig_refine}
\end{figure}

As pointed out in (\ref{eq_prob}),  the softmax layer generates the estimated probability of scene classification. Let's denote $\bm{p_s} \in \mathbb{R}^{1 \times M_s}=[p_s^1,\cdots,p_s^{M_s}]$ as the probability vector. Similarly, the probability of semantic segmentation is denoted as $\bm{p_o} \in \mathbb{R}^{ H \times W \times M_s}$. Then the refinement process can be represented as follow:
\begin{eqnarray}
\bm{p_{so}} = \bm{p_s} \times \bm{W_{so}} \\
\bm{\widetilde{p}_o} = \bm{p_{so}} \otimes \bm{p_o}
\end{eqnarray}
where $\bm{W_{so}} \in \mathbb{R}^{M_s \times M_o}$ is the refinement matrix learned from training data, $\bm{p_{so}}$ represents the prior probability of objects learned from estimated scene classes, which is propagated to $\bm{p_o}$ through multiplication with broadcasting (denoted as $\otimes$ in this paper), i.e. broadcast the $M_o$ values in $\bm{p_{so}}$ to each score map in $\bm{p_s}$ respectively. The refinement process is illustrated in Figure~\ref{fig_refine}.

For the refinement matrix $\bm{W_{so}}$, it is constructed based on the scene-object co-occurrence distribution in training dataset. Rather than directly decide the refinement matrix from the object frequency, we propose to construct  $\bm{W_{so}}$ in a way similar to term frequency-inverse document frequency (tf-idf). Inspired by the inverse document frequency term in tf-idf, how important an object is in a scene is also considered. For example, the object classes ``wall'' and ``floor'' are most common ones and almost appear in every scene. These common classes are usually ignored. When we want the robot to finish a certain task such as ``find the bowl in the kitchen'', these common classes are less meaningful in the context of semantic segmentation, while the training process actually pays more attention to these classes because of their large amount of training samples.

Let's first construct the original term frequency matrix $\bm{f}\in \mathbb{R}^{M_s \times M_o}$, where $f_{ij}$ denotes the count of object $j$ occurs in scene $i$. And then the term frequency is normalized as:
\[tf_{ij} = log(1+f_{ij})
\]

In this paper the inverse document frequency is constructed as:
\[idf_j = \frac{N}{\sum_{i=1}^{N}{tf_{ij}}}
\]

Finally, the $w_{ij}$ in weight matrix $\bm{W_{so}}$ is constructed by the multiplication of these two terms with normalization as:
\[w_{ij} = log(1+tf_{ij} \times idf_j)
\]
where $i=1,\cdots,M_s$, $j=1,\cdots,M_o$. If $w_{ij}=0$, we set $w_{ij}=1e^{-2}$ in case that the training dataset cannot exactly represent scene-object occurrences of the test dataset.

\section{Experiments and results}\label{sec_exp}

We first train and validate our SS-CNN on the SUN RBG-D dataset~\cite{Song2015}, which is an indoor dataset with 10335 RGB-D images in all. In~\cite{Song2015}, the benchmark of scene classification is conducted on a subset of the dataset, which is composed of 19 scene classes with more than 80 samples, while the benchmark of semantic segmentation is conducted on the whole dataset with 45 scene classes. A summary of the datasets used in different tasks is given in Table~\ref{table_sundata}, where the split configuration is provided in the toolbox of SUN RGB-D dataset\footnote{http://rgbd.cs.princeton.edu.}. To make a fair comparison, we also validate the scene classification performance on $S_{19}$ and validate the semantic segmentation performance on $S_{45}$. For both cases, SS-CNN is trained with only the training images in SUN RGB-D dataset, without other data augmentation.

\begin{table}[tpb]
\caption{Summary of SUN RGB-D dataset.}
\label{table_sundata}
\begin{center}
\begin{tabular}{lcccc}
Task 					&	Dataset		&	\#Train 		& 	\#Test & \#All  \\
\hline
Scene classification 	&	$S_{19}$	 &	4845	& 4659 		& 9504 		\\
Semantic segmentation	&   $S_{45}$	 & 	5285 	& 5050		& 10335		\\
\end{tabular}
\end{center}
\end{table}


On top of the model trained and validated on SUN RGB-D dataset, we experimentally test the performance of SS-CNN on a set of test images collected in a building of our university using a mobile robot.

\subsection{Experimental setup}\label{sec_setup}
During the training process, we resize both input images and semantic segmentation ground truth to $210\times 158$ for computation efficiency. Let's denote the resized image datasets as $\hat{S}_{19}$ and $\hat{S}_{45}$  respectively.

To predict the pixel-wise labels in the semantic segmentation branch, we construct our SS-CNN based on a slightly modified Alexnet. The receptive field of the original Alexnet is $224 \times 224$, with pixel stride $32$. Intuitively, large stride leads to coarse semantic segmentation results. Smaller stride is obviously required for semantic segmentation in our work since the image size we use is $210 \times 158$. In~\cite{long2014fully}, the author implemented a fusion technique named ``deep jet'' for finer segmentation results. Instead of fusing results from multiple layer such as using ``deep jet'', we choose to slightly modify the configuration of Alexnet to directly get a network with stride $16$ and receptive field $81 \times 81$. The rationale is this paper focuses on validating the effectiveness of semantic regularization on scene classification rather than obtaining a finer semantic segmentation. Besides, the length of \texttt{fc7} is reduced to 512 while the original length in Alexnet is 4096, which is also illustrated in Figure~\ref{fig_framework}. We believe the performance of SS-CNN can be further improved with higher resolution of training images and more parameters in \texttt{fc7}.

Our network is implemented on Caffe~\cite{jia2014caffe}, a popular deep learning framework. For model learning, we use stochastic gradient descent with momentum to train the randomly initialized network, and the size of each minibatch is 20. The learning rate is fixed as $10^{-4}$ during the training process, and the momentum is fixed as $0.9$. Similar to the common configuration in training deep neural networks, we use a weight decay of $5^{-4}$, and double the learning rate of biases. We also employed dropout in the fully connected layers. We are planning to release our model in the near future.

\subsection{Evaluation of semantic regularization}

To evaluate the effectiveness of our semantic regularization, we first make a comparison between our SS-CNN-R$n$ and the basic Alexnet, where layer 1 to layer $n$ in SS-CNN-R$n$ are regularized by semantic segmentation cost as introduced in Section~\ref{sec_model1}. Both SS-CNN-R$n$ and the original Alexnet are trained with the same training data in $\hat{S}_{19}$, and only RGB images are considered in this evaluation experiment.
\begin{figure}[tpb]
\centering
\includegraphics[height=4cm]{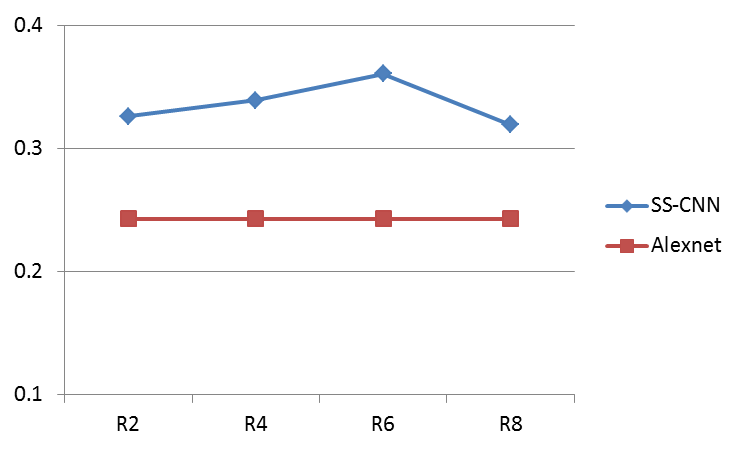}
\caption{Comparison of SS-CNN-R$n$ on scene classification with $n=2,4,6,8$. The performance on the original Alexnet is given as a baseline. }
\label{fig_chart}
\end{figure}

The models we compared include SS-CNN-R2, R4, R6 and R8. Comparison result is shown in Figure~\ref{fig_chart}, which demonstrates SS-CNN-R$n$ considerably outperforms the original Alexnet for each $n$. It reveals the generalization ability on scene classification is significantly improved with the regularization of semantic segmentation.

\begin{figure}[tpb]
\centering
\subfigure[Alexnet]{
\includegraphics[height=3cm]{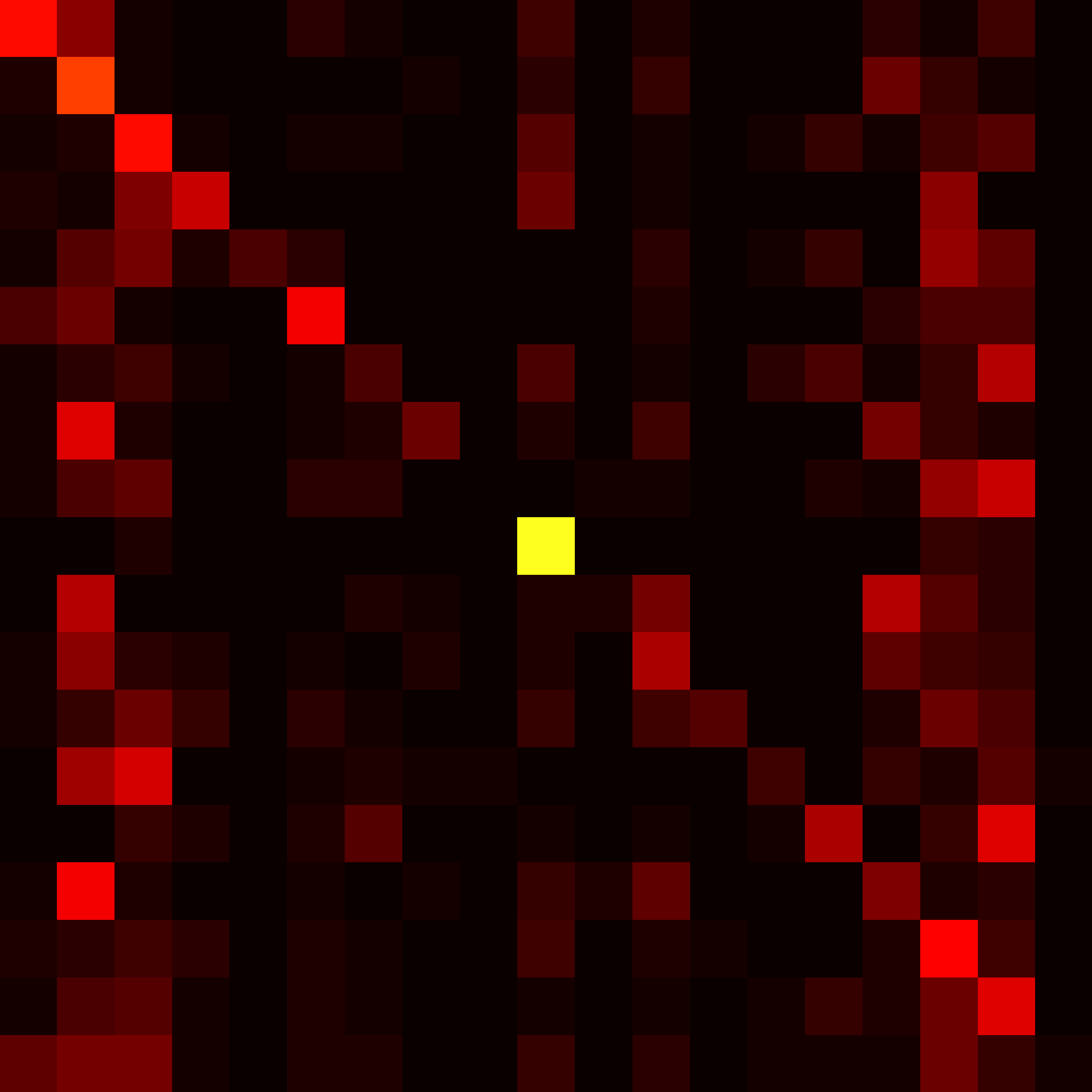}}
\subfigure[SS-CNN-R6]{
\includegraphics[height=3cm]{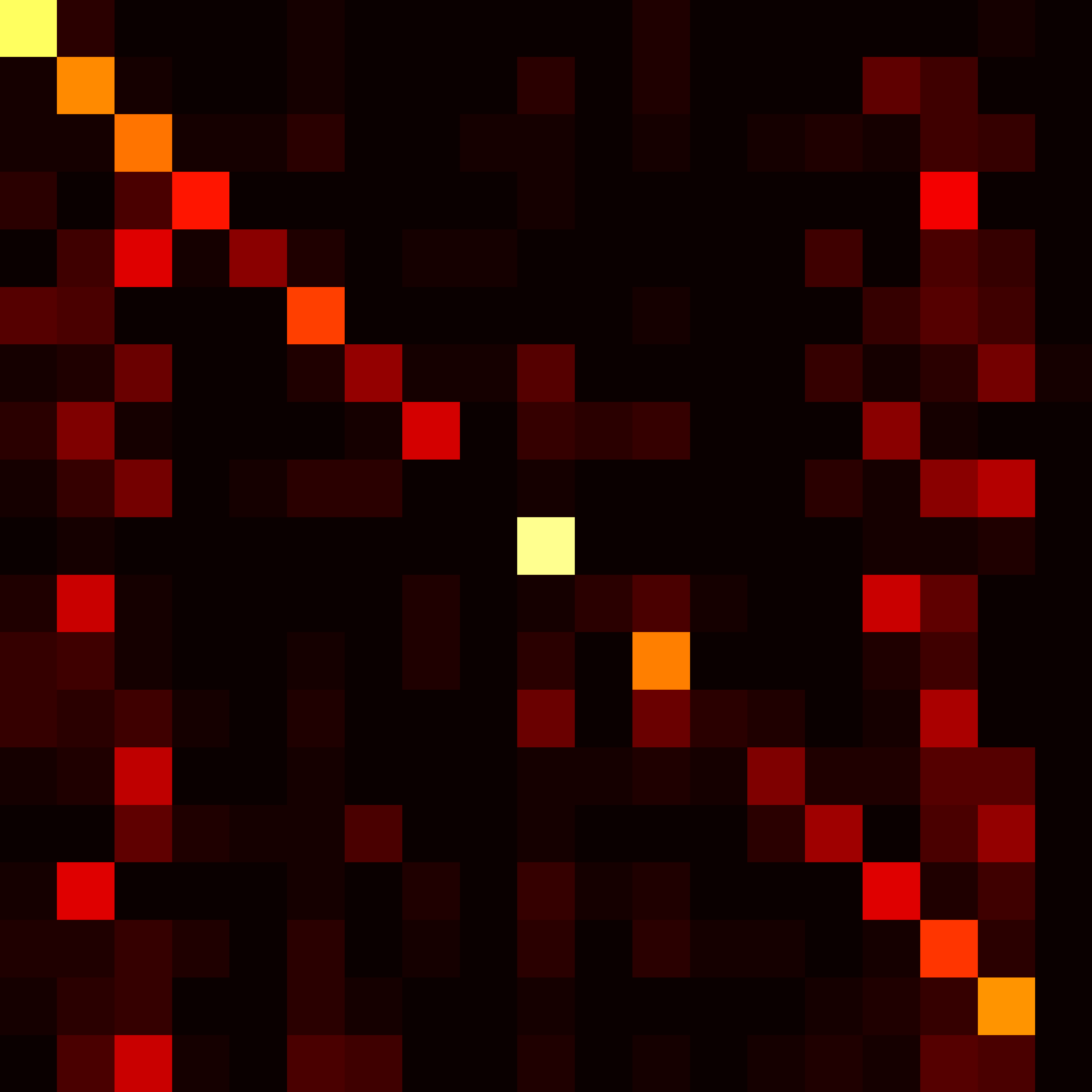}}
\subfigure{
\includegraphics[height=2.98cm]{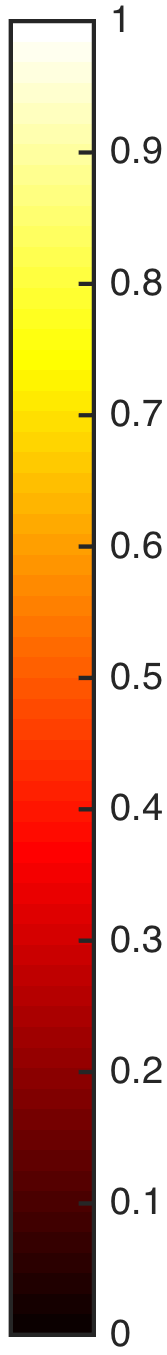}}
\caption{Confusion matrices of Alexnet and SS-CNN-R6, both are trained with only training images in SUN RGB-D dataset.}
\label{fig_confmat}
\end{figure}

By analyzing the influence of $n$ in SS-CNN-R$n$, we can further gain insights in how the generalization ability is enhanced with the regularization.   Figure~\ref{fig_chart} shows the performance of SS-CNN-R$n$ experiences slight promotion with increased $n$ from 2 to 6. It can be explained that when $n$ is small, the regularization is added only to the early layers of main branch, which means the low-level features are regularized. As $n$ is increasing, the features being regularized become more abstract, and even object-level features would start to emerge in higher layers with the regularization on semantic segmentation. However, it can be seen that the performance of SS-CNN-R8 has an apparent drop. One possible reason is that SS-CNN-R8 directly classifies the scene based on semantic segmentation results, in which the performance of scene classification would suffer from the misclassification of semantic segmentation. For better illustration, the confusion matrices of our best model SS-CNN-R6 and Alexnet is given in Figure~\ref{fig_confmat}, which demonstrate the scene classification result is considerably improved with the semantic regularization.

\subsection{Validation on scene classification}

In this section, we make comparison between our SS-CNN and the benchmark methods of scene classification on $S_{19}$~\cite{Song2015}, with 4845 training samples and 4659 validation samples as shown in Table~\ref{table_sundata}. It is to be noted that our SS-CNN is trained and evaluated on the resized $\hat{S}_{19}$, and it does not harm the fairness in the task of scene classification. The introduction of our comparing methods is given as follow:
\begin{itemize}
\item GIST~\cite{oliva2001modeling} + SVM. GIST is a famous descriptor for modeling a scene image, which summarizes the gradient information of a given image.  An RBF kernel SVM is employed for classification.
\item Place-CNN~\cite{zhou2014learning} + SVM. As introduced in Section~\ref{sec_related}, Place-CNN is pre-trained on 2.5 million scene images using Alexnet. Because of its pre-trained structure, Place-CNN is usually employed as a feature extraction method in scene classification applications and an additional classifier is required for classification. In~\cite{Song2015}, both Linear SVM and RBF Kernal SVM are considered to train and classify the Place-CNN features extracted from $S_{19}$, and the later achieves the state-of-the-art performance.
\item Alexnet. As both Place-CNN and SS-CNN are based on Alexnet, the performance of the original Alexnet is also evaluated as a baseline. The Alexnet we implement is trained with only training images in $\hat{S}_{19}$ from randomly initialized weights. Dislike the separate feature extraction and classification required in Place-CNN model, the Alexnet we trained directly classifies the scene classes using the softmax classifier within the network architecture.
\item SS-CNN. As suggested in Figure~\ref{fig_chart}, SS-CNN-R6 is the best configuration and thus is employed in this comparison. Our SS-CNN-R6 is also trained with only training images in $\hat{S}_{19}$ and the softmax classifier is employed for classification.

\end{itemize}

Comparison results are given in Table~\ref{table_scene}, where the accuracy is calculated as the mean accuracy of 19 scene classes. Table~\ref{table_scene} also shows the results learned from both RGB input and RGB-D input. For the depth information,~\cite{Song2015} adopt the HHA~\cite{gupta2014learning} representation in GIST feature and Place-CNN feature. HHA is composed of horizontal disparity, height above ground, and the angle information. As HHA requires inferring of the ground and the gravity direction, our depth representation in Section~\ref{sec_model1} is a more compact choice with also effective performance as shown in Table~\ref{table_scene}.


It can be seen that Place-CNN gains a considerable promotion with pre-training on 2.5 million scene images compared to the original Alexnet trained with only SUN RGB-D dataset. For SS-CNN-R6 which is also trained on SUN RGB-D dataset, it achieves superior results taking advantage of the regularization on semantic segmentation, which is slightly better than the Place-CNN with our RGB-D input. The results further validate our hypothesis that the generalization ability of deep neural network could be enhanced by involving object-level information.


\begin{table}[tpb]
\caption{Scene classification comparison.}
\label{table_scene}
\begin{center}
\begin{tabular}{llc}
Model 		& 	Input 					& Acc (\%)  \\
\hline
GIST + 									& RGB 				& 19.7          \\
RBF Kernel SVM~\cite{Song2015}			& RGB + D 	& 23.0			\\
\hline
Place-CNN +								& RGB 					& 35.6          \\
Linear SVM~\cite{Song2015}				& RGB + D 	& 37.2			\\
\hline
Place-CNN + 							& RGB 					& 38.1	        \\
RBF Kernel SVM~\cite{Song2015}			& RGB + D 	& 39.0			\\
\hline
\multirow{2}{*}{Alexnet}				& RGB & 24.3\\
										& RGB + D & 30.7\\
\hline
\multirow{2}{*}{SS-CNN-R6}	& RGB 					& 36.1		\\
							& RGB + D	& \textbf{41.3}			\\
\end{tabular}
\end{center}
\end{table}

\subsection{Validation on semantic segmentation and its refinement}

We also evaluate the performance of the semantic segmentation, the regularizer, and its refined results. The dataset we use is $S_{45}$ as shown in Table~\ref{table_sundata}, which has 37 object classes. The comparing results are shown in Table~\ref{table_seg}, the accuracy is calculated as the mean accuracy of all 37 objects.

In Tabel~\ref{table_seg}, we first compare the performances of SS-CNN-R6 on $\hat{S}_{45}$, i.e. the resized dataset. Results show that depth information significantly promotes the mean accuracy of semantic segmentation. Then it is further refined to increase the mean accuracy.
In particular, the accuracies on ``chair'', ``ceiling'' and ``bookshelf'' are significantly promoted with refinement.


To make a fair comparison to the benchmark methods mentioned in~\cite{Song2015}, our predicted results on $\hat{S}_{45}$ is directly resized to $S_{45}$, which slightly effects the mean accuracy. The comparing methods are listed as follow:
\begin{itemize}
\item Nearest neighbor. A nonparameteric method, \cite{Song2015} first extracts features using the trained Place-CNN to represent each image, and the test image directly takes the ground truth of the nearest neighbor in feature space as its segmentation label.
\item SIFT Flow~\cite{liu2009nonparametric}. Also a nonparameteric method which takes the SIFT flow matching algorithm to search the match images from dataset with available semantic segmentation.
\item Kernel Descriptors (KDES)~\cite{ren2012rgb}. A state-of-the-art method which encodes the input with kernel descriptors and the contextual information is considered with superpixel MRF and segmentation tree.
\end{itemize}

\begin{figure*}[tpb]
\centering
\subfigure{
\includegraphics[height=2.55cm]{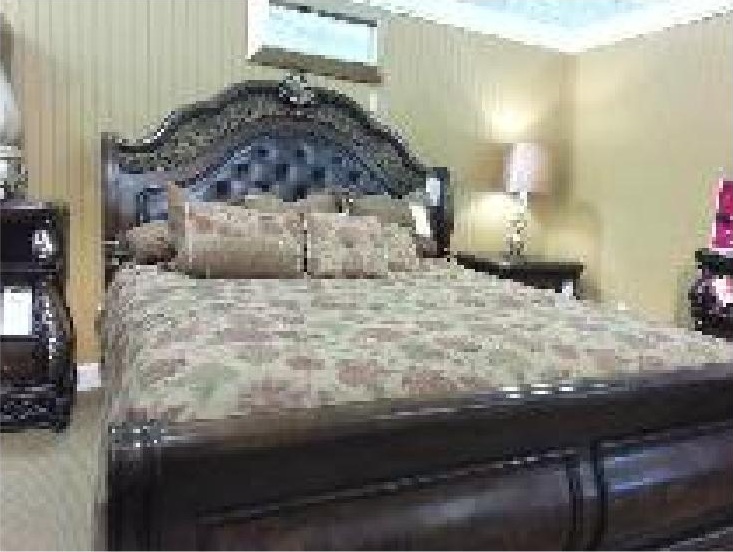}}
\subfigure{
\includegraphics[height=2.55cm]{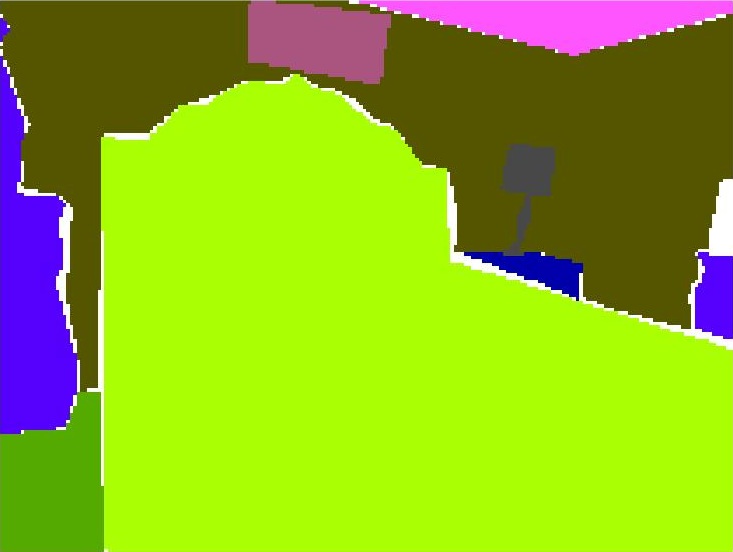}}
\subfigure{
\includegraphics[height=2.55cm]{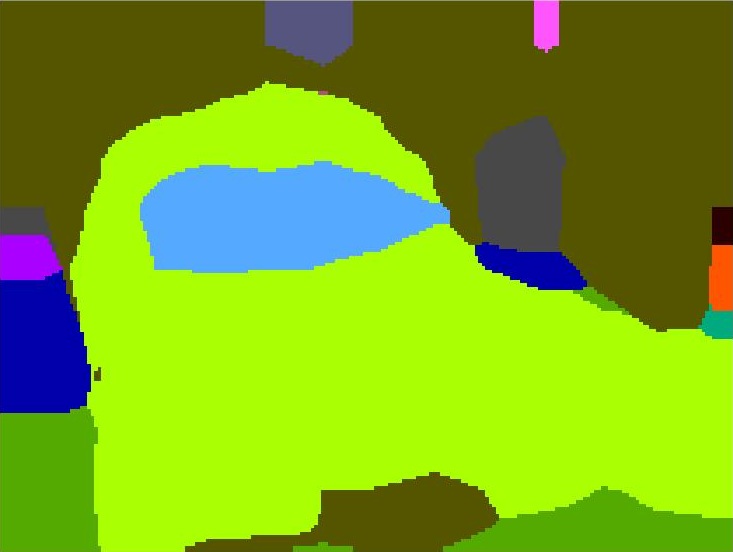}}
\subfigure{
\includegraphics[height=2.55cm]{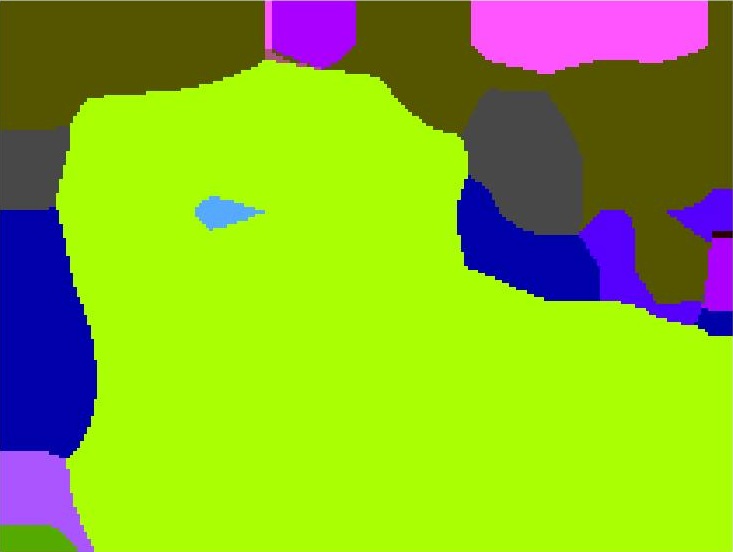}}
\subfigure{
\includegraphics[height=2.55cm]{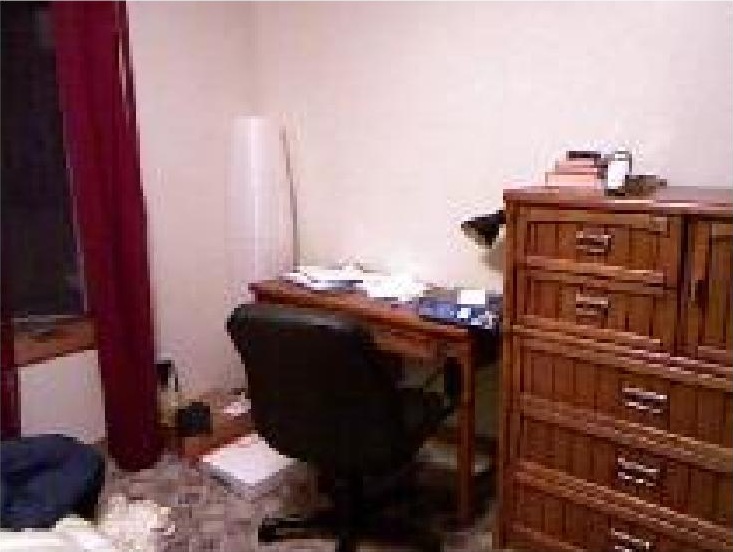}}
\subfigure{
\includegraphics[height=2.55cm]{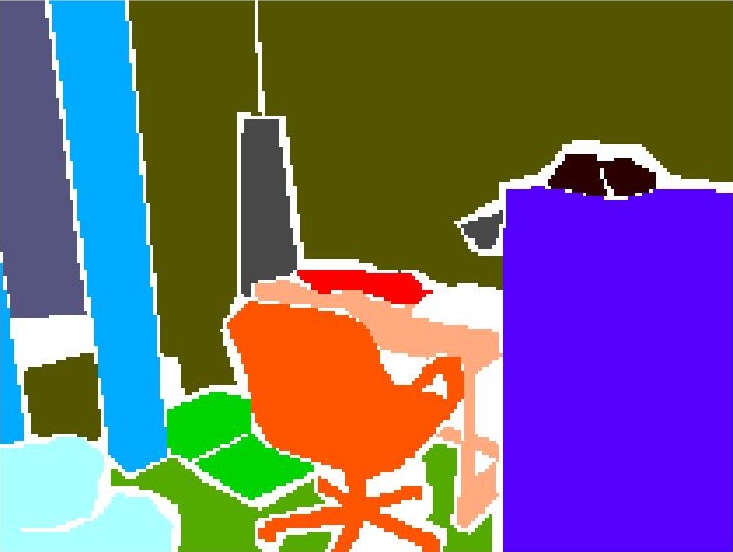}}
\subfigure{
\includegraphics[height=2.55cm]{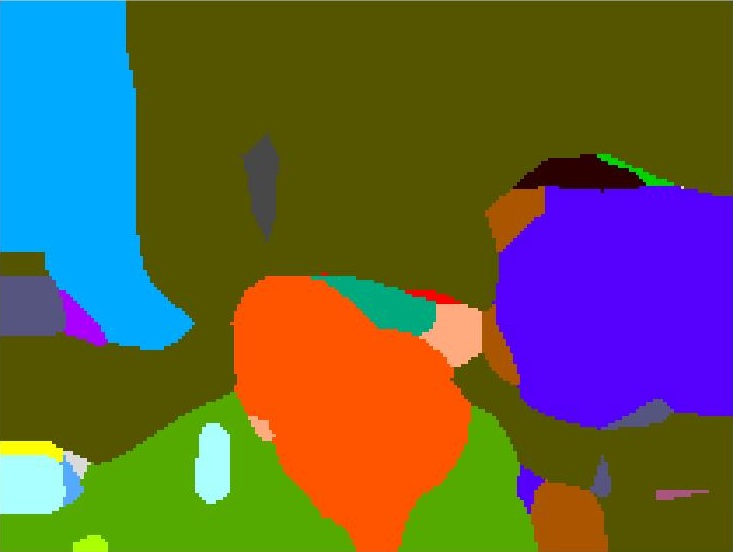}}
\subfigure{
\includegraphics[height=2.55cm]{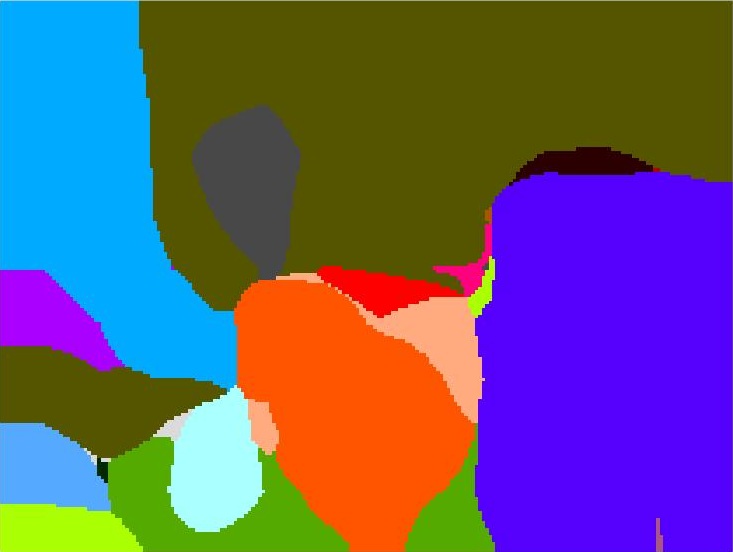}}

\caption{Illustration of semantic segmentation and its refinement. From left to right: RGB input, ground truth of semantic segmentation, predicted results of SS-CNN-R6, refined predicted results of SS-CNN-R6. White color in the ground truth images denotes the background or confusing region and not considered either in training nor test. It can be seen that refinement not only plays the role of smoothing, but also ``strengthens'' some specific objects in the corresponding scene.}
\label{fig_02074}
\end{figure*}

As can be seen in Table~\ref{table_seg}, on dataset $S_{45}$, we also achieve the state-of-the-art performance on semantic segmentation with the SS-CNN-R6. We illustrate some examples of our predicted semantic segmentation labels with their refined results in Figure~\ref{fig_02074}.

\begin{table}[tpb]
\caption{Semantic segmentation comparison.}
\label{table_seg}
\begin{center}
\begin{tabular}{cllc}
Dataset &	Model 		& 	Input 				 & Acc (\%)  \\
\hline
\multirow{3}{*}{$\hat{S}_{45}$} & \multirow{3}{*}{SS-CNN-R6} 	& RGB 	& 27.77	\\
						  	&& RGB + D 			& 37.03 \\
						  	&& RGB + D refined 	&  \textbf{41.76} \\


\hline
\multirow{4}{*}{$S_{45}$}
&NN~\cite{Song2015}			
							& RGB + D 	& 8.97			\\
&SIFT Flow~\cite{Song2015}	
							& RGB + D  	& 10.05			\\
&KDES~\cite{Song2015}		& RGB + D   & 36.33	\\
&SS-CNN-R6 & RGB + D refined & \textbf{40.66}\\

\end{tabular}
\end{center}
\end{table}

\begin{table}[tpb]
\caption{Experimental validation results.}
\label{table_uts}
\begin{center}
\begin{tabular}{lcc}
Class &	\#Sample		& 	Acc (\%)  \\
\hline
Computer room	 & 	41 	& 19.5\\
Conference room	 &	29	& 13.8 \\
Corridor		 &  38	& 47.4 \\
Kitchen 		 &  14	& 35.7 \\
Office			 &  94  & 63.8 \\
Rest space		 &  14  & 57.1 \\
\hline
All 			 &  230 & 39.6 \\
\end{tabular}
\end{center}
\end{table}

\subsection{Experimental validation}

The experiments on the publicly available SUN RGB-D dataset demonstrates the effectiveness of our SS-CNN. To further validate the performance of SS-CNN in robotics related application, we conducted an experiment using our mobile robot. The robot moved around in one of our university buildings and collected 230 RGB-D images with an on-board Kinect V2, belonging to 6 scene classes.

For scene classification, we use the SS-CNN-R6 training on SUN RGB-D dataset to predict the scene classes in the collected images without retraining the network with images in the new environment. To adapt to the SS-CNN-R6, each collected image is also represented as catenation of RGB image, depth image and normal vector image. The predicted results are given in Table~\ref{table_uts}, where the mean accuracy of all these 6 classes are given at the bottom row. Figure~\ref{robot} gives some example RGB images with their predicted labels. It is to be noted that some images in this dataset is challenging even for humans since the boundary between some scene classes are not very clear. The last row in Figure~\ref{robot} gives two examples in this situation, the ground truth of these two images are ``computer room'' and ``rest space'' respectively, while they are denoted with ``office'' and ``discussion area''.

As can be seen from Table~\ref{table_uts}, the predicted results are in similar order to the validation results on SUN RGB-D in the completely new environment, which further demonstrates the generalization ability of our SS-CNN. Therefore, our SS-CNN has the potential to be implemented in real robotics applications without further training.

\section{Conclusion}\label{sec_conclu}
In this paper, we address the scene classification problem using deep learning methods with a much smaller amount of training images, by regularizing deep architecture with semantic segmentation. Experimental results validate the effectiveness of the regularization as SS-CNN achieved the state-of-the-art results on both scene classification and semantic segmentation on the publicly available SUN RGB-D dataset. Further experiments on our robot demonstrates the generalization ability of the proposed approach. For the future work, we would like to investigate the potential possibility in both horizontal and vertical dimensions, which means to couple more relevant tasks, and to find better architecture to incorporate the relations between these tasks.

\bibliographystyle{ieeetr}
\bibliography{scene_seg}

\end{document}